%% file: main.tex
\documentclass[runningheads]{llncs}
\usepackage{graphicx}
\usepackage{comment}
\usepackage{amsmath,amssymb} 
\usepackage{color}
\usepackage{epsfig}
\usepackage{booktabs}
\usepackage{xcolor, colortbl}
\usepackage{tabularx}
\usepackage{multirow}

\usepackage{url}
\usepackage{color}
\usepackage{subfigure}
\usepackage{mwe}
\usepackage{booktabs,arydshln}
\usepackage{bbm}

\usepackage[pagebackref=true,breaklinks=true,letterpaper=true,colorlinks,bookmarks=false]{hyperref}
\usepackage{breakcites}
\hypersetup{
    colorlinks   = true,
    citecolor    = green
}
\hypersetup{linkcolor=red}

\newcommand\etal{\textit{et al.}}
\newcommand\eg{\textit{e.g.}}
\newcommand\ie{\textit{i.e.}}

\begin{document}
\pagestyle{headings}
\mainmatter
\def\ECCVSubNumber{3935}  

\title{Learning to Optimize Domain Specific Normalization for Domain Generalization} 

\titlerunning{Learning to Optimize Domain Specific Normalization}
%
\author{Seonguk Seo\inst{1} \and
Yumin Suh\inst{2} \and
Dongwan Kim\inst{1} \and \\
Geeho Kim\inst{1} \and
Jongwoo Han\inst{3} \and
Bohyung Han\inst{1}
}
\authorrunning{S. Seo et al.}
%
\institute{Seoul National University \and
NEC Laboratories America\and
LG Electronics
}
\maketitle

\begin{abstract}
We propose a simple but effective multi-source domain generalization technique based on deep neural networks by incorporating optimized normalization layers that are specific to individual domains.
Our approach employs multiple normalization methods while learning separate affine parameters per domain.
For each domain, the activations are normalized by a weighted average of multiple normalization statistics.
The normalization statistics are kept track of separately for each normalization type if necessary.
Specifically, we employ batch and instance normalizations in our implementation to identify the best combination of these two normalization methods in each domain.
The optimized normalization layers are effective to enhance the generalizability of the learned model.
We demonstrate the state-of-the-art accuracy of our algorithm in the standard domain generalization benchmarks, as well as viability to further tasks such as multi-source domain adaptation and domain generalization in the presence of label noise.
\keywords{Domain generalization}
\end{abstract}

\input{sections/intro.tex}

\input{sections/rel_work.tex}
\input{sections/method.tex}
\input{sections/exp.tex}

\input{sections/conclusion.tex}

%
%
\bibliographystyle{splncs04}
\bibliography{egbib}
\end{document}

%% file: sections/intro.tex

\section{Introduction}

Domain generalization aims to learn generic feature representations agnostic to domains and make trained models perform well in completely new domains.
To achieve this challenging goal, one needs to train models that can capture useful information observed commonly in multiple domains and recognize semantically related but visually inconsistent examples effectively.
Many real-world problems have similar objectives so this task can be widely used in various practical applications.
Domain generalization is closely related to unsupervised domain adaptation but there is a critical difference regarding the availability of target domain data; contrary to unsupervised domain adaptation, domain generalization cannot access any examples in target domain during training but is still required to capture transferable information across domains.
Due to this constraint, the domain generalization problem is typically considered to be more challenging, so multiple source domains are usually involved to make the problem more feasible.

Domain generalization techniques are classified into several groups depending on their approaches.
Some algorithms define novel loss functions to learn domain-agnostic representations~\cite{muandet2013domain,motiian2017unified,li2018domain,dou2019domain} while others are more interested in designing deep neural network architectures to achieve similar goals~\cite{pacs,dinnocente2018domain,mancini2018best}.
The algorithms based on meta-learning have been proposed under the assumption that there exists a held-out validation set~\cite{balaji2018metareg,li2019episodic,li2018learning}.

Our algorithm belongs to the second category, \ie~network architecture design methods.
In particular, we are interested in exploiting normalization layers in deep neural networks to handle the domain generalization task.
A na\"ive approach would be to train a single deep neural network with batch normalization using all training examples regardless of their domain memberships.
This method works fairly well partly because batch normalization regularizes feature representations from heterogeneous domains and the trained model is often capable of adapting to unseen domains.
However, the benefit of batch normalization is limited when domain shift is significant, and we are often required to remove domain-specific styles for better generalization.
Instance normalization~\cite{in} turns out to be an effective scheme for the goal and incorporating both  batch and instance normalization techniques further improves accuracy by a data-driven balancing of two normalization methods~\cite{bin,ibn}.
Our approach also employs the two normalizations but proposes a more sophisticated algorithm designed for domain generalization. 

\begin{figure*}[t]
\centering
	\includegraphics[width=0.99\linewidth]{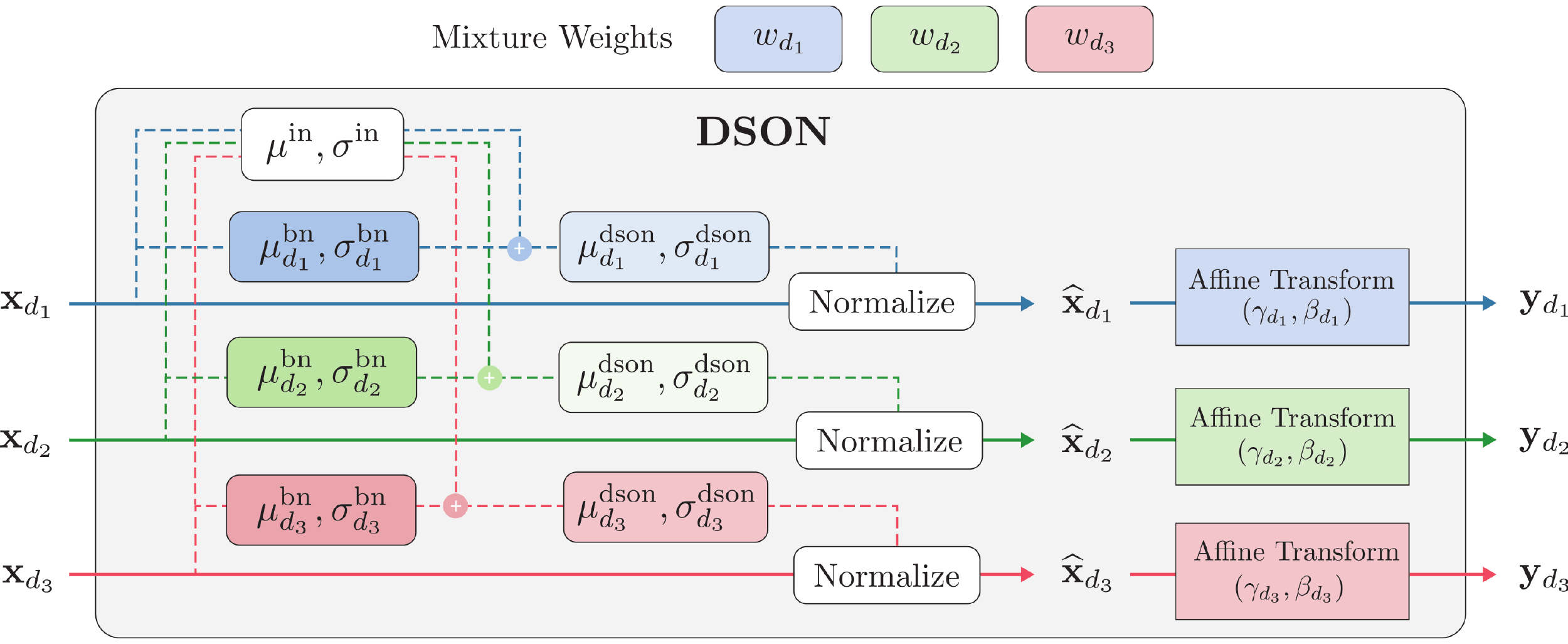}
   \caption{Illustration of Domain Specific Optimized Normalization (DSON). Each domain maintains domain-specific batch normalization statistics and affine parameters, as well as mixture weights.
}
    \label{fig:DSON}
\end{figure*}

We explore domain-specific normalizations to learn representations that are both domain-agnostic and semantically discriminative by discarding domain-specific ones.
The goal of our algorithm is to optimize the combination of normalization techniques in each domain while different domains learn separate parameters for the mixture of normalizations.
The intuition behind this approach is that we can learn domain-invariant representations by controlling types of normalization and parameters in normalization layers.
Note that all other parameters, including the ones in convolutional layers, are shared across domains. 
Although our approach is somewhat similar to \cite{sn} in that the optimal mixing weights between normalization types are learned, 
we emphasize that the motivations are different; \cite{sn} aims for a differentiable normalization for universal tasks while we set our sights on how to remove style information without losing semantics to generalize on unseen domains.
In addition, our domain-specific properties---learning normalization parameters, batch statistics and mixture weights for each domain separately--- makes it unique and more effective to construct domain-agnostic representations, thereby outperforming all the established normalization techniques.
We illustrate the main idea of our approach in Fig.~\ref{fig:DSON}.

Our contributions are as follows:
\begin{itemize}
\item[$\bullet$] Our approach leverages instance normalization to optimize the trade-off between cross-category variance and domain invariance, which is desirable for domain generalization in unseen domains.
\item[$\bullet$] We propose a simple but effective domain generalization technique combining heterogeneous normalization methods specific to individual domains, which facilitates the extraction of domain-agnostic feature representations by removing domain-specific information effectively.
\item[$\bullet$] The proposed algorithm achieves the state-of-the-art accuracy in multiple standard benchmark datasets and outperforms all established normalization methods.
\end{itemize}

%% file: sections/rel_work.tex

\section{Related Work}
\label{sec:related}

This section discusses existing domain generalization approaches and reviews two related problems, multi-source domain adaptation and normalization techniques in deep neural networks.

\subsection{Domain Generalization}
Domain generalization algorithms learn domain-invariant representations given input examples regardless of their domain memberships.
Since target domain information is not available at training time, they typically rely on multiple source domains to extract knowledge applicable to any unseen domain.
The existing domain generalization approaches can be roughly categorized into three classes.
The first group of methods proposes novel loss functions that encourage learned representations to generalize well to new domains.
Muandet~\etal~\cite{muandet2013domain} propose domain-invariant component analysis, which generates invariant feature representation via dimensionality reduction.
A few recent works~\cite{motiian2017unified,li2018domain,dou2019domain} also attempt to learn a shared embedding space appropriate for semantic matching across domains.
Another kind of approach tackles the domain generalization task by manipulating deep neural network architectures.
Domain-specific information is handled by designated modules within deep neural networks \cite{pacs,dinnocente2018domain,matsuura2020domain} while \cite{mancini2018best} proposes a soft model selection technique to obtain generalized representations.
Recently, meta-learning based techniques start to be used to solve domain generalization problems.
MLDG~\cite{li2018learning} extends MAML~\cite{finn2017model} to domain generalization task.
Balaji~\etal~\cite{balaji2018metareg} points out the limitation of \cite{li2018learning} and proposes a regularizer to address domain generalization in a meta-learning framework directly.
Also, \cite{li2019episodic} presents an episodic training technique appropriate for domain generalization.
Note that, to the best of our knowledge, none of the existing methods exploit normalization types and their optimization for domain generalization.

\subsection{Multi-Source Domain Adaptation}
Multi-source domain adaptation can be considered as the middle-ground between domain adaptation and generalization, where data from multiple source domains are used for training in addition to examples in an unlabeled target domain. 
Although unsupervised domain adaptation is a very popular problem, its multi-source version is relatively less investigated. 
Zhao~\etal~\cite{zhao2018adversarial} propose to learn features that are invariant to multiple domain shifts through adversarial training, and Guo~\etal~\cite{guo2018multi} use a mixture-of-experts approach by modeling the inter-domain relationships between source and target domains.
A recent work using domain-specific batch normalization (DSBN)~\cite{dsbn} has shown competitive performance in multi-source domain adaptation by aligning the representations in heterogeneous domains to a single common feature space.

\subsection{Normalization in Neural Networks}
Normalization techniques in deep neural networks are originally designed for regularizing trained models and improving their generalization performance.
Various normalization techniques~\cite{bn,in,ln,spectralnorm,wn,gn,bin,sn,ssn,dsbn} have been studied actively in recent years. 
The most popular technique is batch normalization (BN)~\cite{bn}, which normalizes activations over individual channels using data in a mini-batch while instance normalization (IN)~\cite{in} performs the same operation per instance instead of mini-batch.
In general, IN is effective to remove instance-specific characteristics (\eg~style in an image) and adding IN makes a trained model focus on instance-invariant information and increases generalization capability of the model to an unseen domain.
Other normalizations such as layer normalization (LN)~\cite{ln} and group normalization (GN)~\cite{gn} have the same concept while weight normalization~\cite{wn} and spectral normalization~\cite{spectralnorm} normalize weights over parameter space. 

Recently, batch-instance normalization (BIN)~\cite{bin}, switchable normalization (SN)~\cite{sn}, and sparse switchable normalization (SSN)~\cite{ssn} employ the combinations of multiple normalization types to maximize the benefit.
Note that BIN considers batch and instance normalizations while SN uses LN additionally.
On the other hand, DSBN~\cite{dsbn} adopts separate batch normalization layers for each domain to deal with domain shift and generate domain-invariant representations.

%% file: sections/method.tex

\section{Domain-Specific Optimized Normalization for Domain Generalization}
\label{sec:method}
This section describes our main algorithm called domain-specific optimized normalization (DSON) in details and also presents how the proposed method is employed to solve domain generalization problems.

\subsection{Overview}

Domain generalization aims to learn a domain-agnostic model that can be applied to an unseen domain by leveraging multiple source domains.
Consider a set of training examples $\mathcal{X}_s$ with its corresponding label set $\mathcal{Y}_s$ in a source domain $s$.
Our goal is to train a classifier using the data in multiple source domains $\{\mathcal{X}_s\}_{s=1}^{S}$ to correctly classify an image $\mathbf{x}_t \in \mathcal{X}_t$ in a target domain $t$, which is unavailable during training.

In our approach, we aim to learn a joint embedding space across all source domains, which is expected to be valid in target domains as well.
To this end, we train domain-invariant classifiers from each of the source domains and ensemble their predictions.
To embed each example onto a domain-invariant feature space, we employ domain-specific normalization, which is to be described in the following sections.

Our classification network consists of a set of feature extractors $\{F_s\}_{s=1}^S$ and a single fully connected layer $D$.
Specifically, the feature extractors share all parameters across domains except for the ones in the normalization layers.
For each source domain $s$, loss function is defined as
\begin{align}
\mathcal{L}_C(\mathcal{X}_s, \mathcal{Y}_s) = \frac{1}{|\mathcal{X}_s|} \sum_{x \in \mathcal{X}_s, y\in \mathcal{Y}_s}{\ell(y, D(F_s(x))},
\end{align}
where $\ell(\cdot, \cdot)$ is the cross-entropy loss.
All the parameters are jointly optimized to minimize the sum of classification losses of source domains:
\begin{equation}
\label{eq:loss}
\mathcal{L} = \sum_{s=1}^{S}{\mathcal{L}_C(\mathcal{X}_s, \mathcal{Y}_s)}.
\end{equation}
Our domain-specific deep neural network model is obtained by minimizing the total loss $\mathcal{L}$.
{\color{black}
To facilitate generalization, in the validation phase, we follow the leave-one-domain-out validation strategy proposed in \cite{dinnocente2018domain};
the label of a validation example from domain $s$ is predicted by averaging predictions from all domain-specific classifiers, except for the one with domain $s$.
}

\begin{figure*}
\begin{center}
	\subfigure[Input]{\fbox{\includegraphics[width=0.3\linewidth]{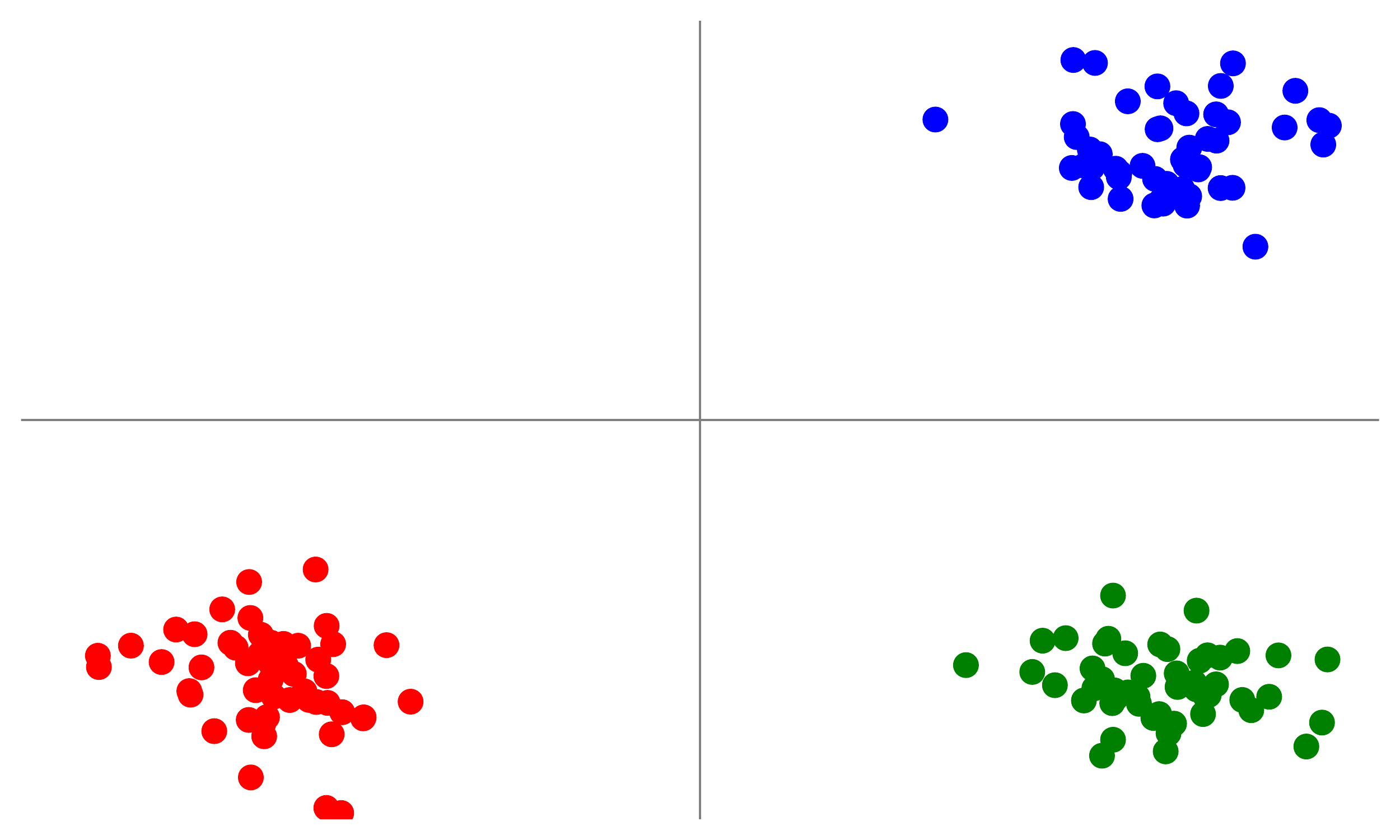}}}
	\subfigure[Batch Normalization]{\fbox{\includegraphics[width=0.3\linewidth]{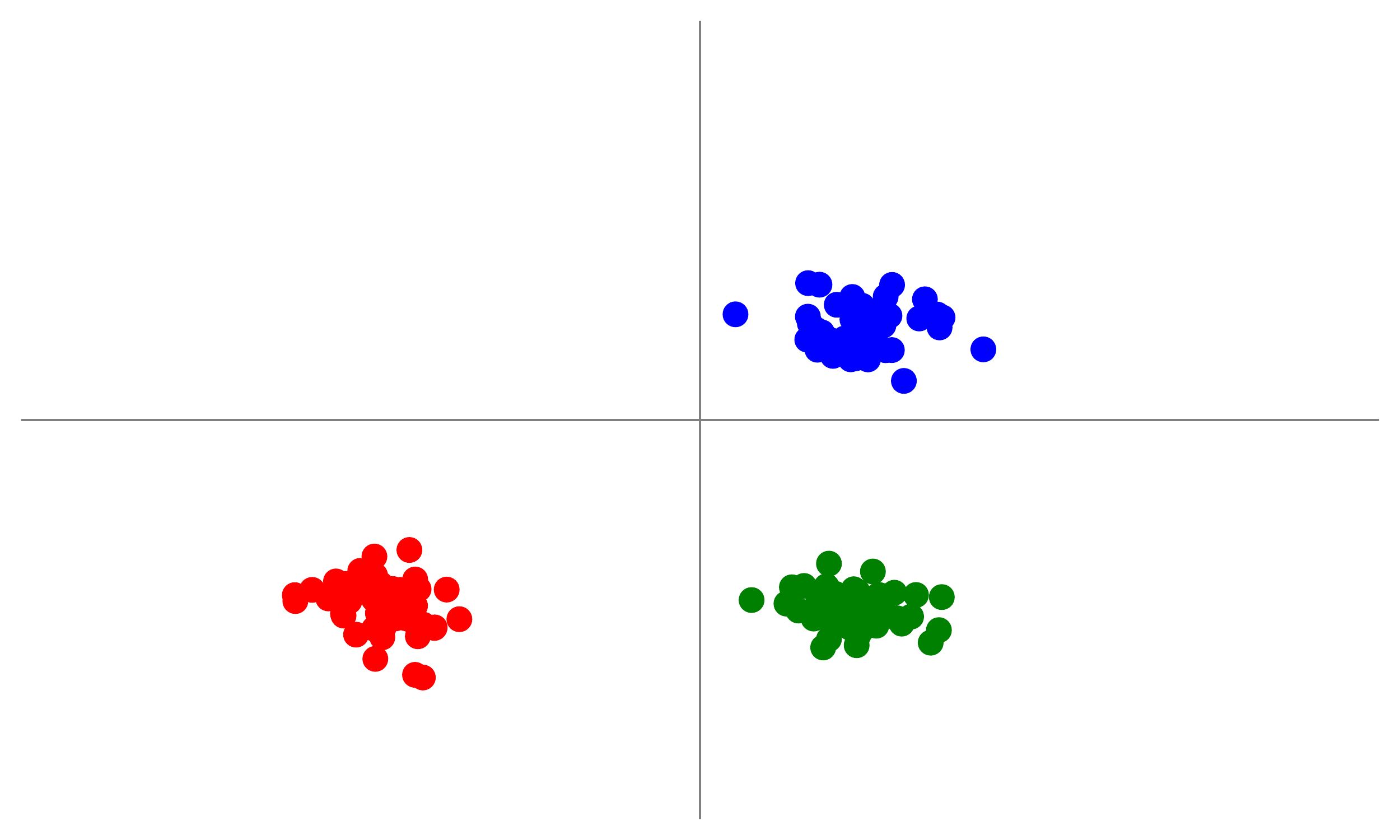}}}
	\subfigure[Instance Normalization]{\fbox{\includegraphics[width=0.3\linewidth]{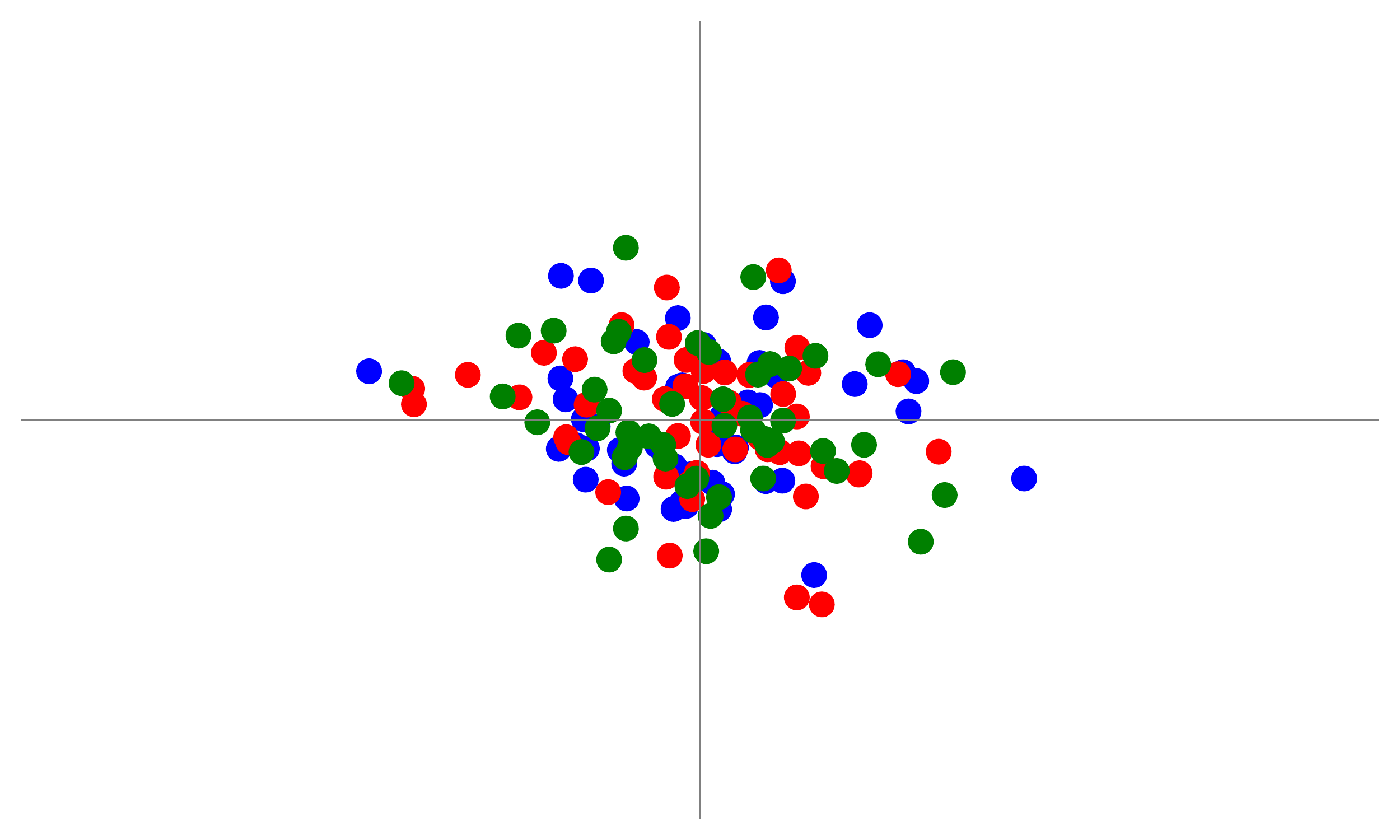}}}
	\caption{Comparing feature distributions of three classes, where color represents the class label and each dot represents a feature map with two channels. where each axis corresponds to one channel. For given (a) input activations, (c) instance normalization makes the features less discriminative over classes when compared to (b) batch normalization. Although instance normalization loses discriminability, it makes the normalized representations less overfit to a particular domain and eventually improves the quality of features when combined with batch normalization. (Best viewed in color.)}
	\label{fig:normailization}
\end{center}
\end{figure*}

\subsection{Instance Normalization for Domain Generalization}
\label{sub:batch}

Normalization techniques~\cite{bn,in,ln,gn} are widely applied in recent network architectures for better optimization and regularization.
Particularly, batch normalization (BN)~\cite{bn} improves performance and generalization ability in training neural networks and has become indispensable in many deep neural networks.
However, when BN is applied to cross-domain scenarios, it may not be optimal~\cite{bilen2017universal}.
Table~\ref{tab:bn} empirically validates our claim about the effects of training BN in domain generalization.
We employ a pretrained ResNet-18 on the ImageNet dataset as the backbone network, and fine-tune it on the PACS dataset with two different settings; fixing the BN parameters and statistics or fine-tuning them.
Although BN generally works well in a variety of vision tasks, it consistently degrades performance when it is trained in the presence of a large domain divergence.
This is because the batch statistics overfit to the particular training domains, resulting in poor generalization performance in unseen domains.
This motivates us to construct a domain-agnostic feature extractor.

\begin{table}[!t]
	\begin{center}
	\caption{Effects of training batch normalization on the PACS dataset using a ResNet-18 architecture.
	Each column shows the performance on the target domain when a network is trained using the remaining domains as sources.
	Fine-tuning BN parameters degrades the generalization performance by overfitting to source domains.
	}
	\label{tab:bn}
	\scalebox{0.95}{
	\setlength\tabcolsep{5pt}
\begin{tabular}{lcccc|c}
\toprule
{} & Art painting & Cartoon & Sketch & Photo & Avg. \\
\hline
BN fixed & \bf{79.25} & \bf{74.61} & \bf{71.52} & \bf{95.99} & \bf{80.34} \\
BN finetuned  & 78.47 & 70.41 & 70.68 & 95.87 & 78.86 \\
\bottomrule
\end{tabular}
}
\end{center}
\end{table}

To achieve our goal, we combine BN with instance normalization (IN) to obtain domain-agnostic features.
Our intuition about the two normalization methods is as follows.
Instance normalization has been widely adopted in many works regarding style transfer due to its ability to perform \textit{style normalization}~\cite{AdaIn2017}.
Inspired by this, we employ IN as a means of reducing the inherent style information in each domain. 
In addition, IN does not depend on mini-batch construction or batch statistics, which can be helpful to extrapolate on unseen domains.
These properties allow the network to learn feature representations that less overfit to a particular domain.
The downside of IN, however, is that it makes the features less discriminative with respect to object categories. This is illustrated in a simplified setting (Fig.~\ref{fig:normailization}), where we represent an instance by a cluster of data points and the corresponding classes by color. Unlike BN, which retains variation across the different classes, IN largely reduces the inter-class variance. To reap the benefits of IN while maintaining good classification performance, we utilize a mixture of IN and BN by optimizing the tradeoff between cross-category variance and domain invariance. More specifically, we fuse IN into all the BN layers of our network by linearly interpolating the means and variances of the two normalization statistics. The combination serves as a regularization method which results in a strong classifier that tends to focus on high-level semantic information but is much less vulnerable to domain shifts.

\subsection{Optimization for Domain-Specific Normalization}
\label{sub:dson}
{\color{black}
Based on the intuitions above, we propose a domain-specific optimized normalization (DSON) for domain generalization.}
Given an example from domain $d$, the proposed domain-specific normalization layer transforms channel-wise whitened activations using affine parameters $\beta_d$ and $\gamma_d$.
{\color{black}Note that whitening is also performed for each domain.}
At each channel, the activations $\mathbf{x}_d \in \mathbb{R}^{H\times W\times N}$ are transformed as
\begin{align}
\text{DSON}_d(\mathbf{x}_d[i, j, n] ; \gamma_d, \beta_d) &= \gamma_d \cdot \hat{\mathbf{x}}_d [i, j, n] + \beta_d,
\end{align}
where the whitening is performed using the domain-specific mean and variance, $\mu_{dn}$ and $\sigma_{dn}^2$,
\begin{align}
\hat{\mathbf{x}}_d[i, j, n] =  \frac{\mathbf{x}_d[i, j, n] - \mu_{dn}}{\sqrt{\sigma_{dn}^{2} + \epsilon}}.
\end{align}

We combine batch normalization (BN) with instance normalization (IN) in a similar manner to \cite{sn} as
\begin{align}
\mu_{dn} &= w_{d}\mu^{\text{bn}}_{d} + (1-w_{d})\mu^{\text{in}}_{n},\\
\sigma_{dn}^2 &= w_{d}\sigma{^{\text{bn}}_d}^2 + (1-w_{d})\sigma{^{\text{in}}_n}^2,
\end{align}
where both are calculated separately in each domain as
\begin{align}
\mu^{\text{bn}}_{d} = \frac{\sum_n \sum_{i,j} \mathbf{x}_d[i,j, n]}{N \cdot H \cdot W}~~~~\text{and}~~~~
\sigma{^{\text{bn}}_{d}}^2 = \frac{\sum_n \sum_{i,j} \left( \mathbf{x}_d[i, j, n] - \mu^{\text{bn}}_{d} \right)^2 }{N \cdot H \cdot W},
\end{align}
and
\begin{align}
\mu^{\text{in}}_{n} = \frac{\sum_{i,j} \mathbf{x}_d[i,j, n]}{H \cdot W}~~~~\text{and}~~~~
\sigma{^{\text{in}}_{n}}^2 = \frac{\sum_{i,j} \left( \mathbf{x}_d[i, j, n] - \mu^{\text{in}}_{n} \right)^2 }{H \cdot W}.
\end{align}
The optimal mixture weight, $w_d$, between BN and IN are trained to minimize the loss in Equation~\ref{eq:loss}.
Note that our domain-specific mixture weights are shared across all layers for each domain, which facilitates to find the optimal point.

\subsection{Inference}

A test example $x$ in a target domain is unknown during training.
Hence, for inference, we feed the example to the feature extractors of all domains.
The final label prediction is given by computing the logits using the fully connected layer $D$, averaging the logits, \ie $\frac{1}{S}\sum_{s=1}^{S}{D(F_s(x))}$, and finally applying a softmax function.

One potential issue in the inference step is whether target domains can rely on the model trained only on source domains.
This is the main challenge in domain generalization, which assumes that reasonably good representations of target domains can be obtained from the information in source domains only. 
{\color{black}
In our algorithm, instance normalization in each domain has the capability to remove domain-specific styles and standardize the representation.}
Since each domain has different characteristics, we learn the relative weights of instance normalization in each domain separately.
Thus, predictions in each domain should be accurate enough even for the data in target domains. Additionally, the accuracy given by aggregating the predictions of multiple networks trained on different source domains should further improve accuracy.

%% file: sections/exp.tex

\section{Experiments}
\label{sec:experiments}

To depict the effectiveness of domain-specific optimized normalization (DSON), we implement it on domain generalization benchmarks and provide an extensive ablation study of the algorithm.

\subsection{Experimental Settings}

\begin{table}[t]
	\begin{center}
		\caption{Comparision with the state-of-the-art domain generalization methods (\%) on the PACS dataset using ResNet-18 and ResNet-50 architectures. Column title indicates the target domain. *All experiments use the ``train'' split for the source domains, except MetaReg~\cite{balaji2018metareg}, which uses both ``train'' and ``validation'' splits.}
		\label{tab:PACS_result}
		\scalebox{0.9}{
			\setlength\tabcolsep{6pt}
			\begin{tabular}{clcccc|c}
				\toprule
				Archictecture & Method & Art painting & Cartoon & Sketch & Photo & Avg. \\ \hline
				\multirow{8}{*}{ResNet-18} & Baseline    & 78.47 & 70.41 & 70.68 & 95.87 & 78.86 \\
				&JiGen~\cite{li2017domain} & 79.42 & 75.25 & 71.35 & {96.03} & 80.51 \\
				&D-SAM~\cite{dinnocente2018domain}	& 77.33 & 72.43 & 77.83 & 95.30 & 80.72\\
				&Epi-FCR~\cite{li2019episodic} & {82.10} & {77.00} & 73.00 & 93.90 & {81.50}\\
				&MetaReg*~\cite{balaji2018metareg}	& 83.70 & {77.20} & 70.30 & 95.50 & {81.70} \\
				&MASF~\cite{dou2019domain} & 80.29 & 77.17 & 71.69 & 94.99 & 81.04 \\
				&MMLD~\cite{matsuura2020domain} & 81.28 & 77.16 & 72.29 & \bf{96.09} & 81.83 \\
				\multirow{6}{*}{ResNet-50} & DSON (Ours)      & \bf{84.67} & \bf{77.65} &\bf{82.23}& {95.87} & \bf{85.11} \\ \hline
				&Baseline   & 80.22 & 78.52 & 76.10 & 95.09 & 82.48 \\
				&MetaReg*~\cite{balaji2018metareg}	& \bf{87.20} & 79.20 & 70.30 & \bf{97.60} & 83.60 \\
				&MASF~\cite{dou2019domain} & 82.89 & 80.49 & 72.29 &  95.01 & 82.67 \\
				&DSON (Ours)      & {87.04} & \bf{80.62} &\bf{82.90}& {95.99} & \bf{86.64} \\
				\bottomrule
		\end{tabular}}
	\end{center}
\end{table}
\begin{table}[t]
	\begin{center}
	\caption{Comparision with the state-of-the-art domain generalization methods (\%) on the Office-Home dataset using a ResNet-18 architecture. Column title indicates the target domain.}
	\label{tab:OH_result}
	\scalebox{0.95}{
	\setlength\tabcolsep{8pt}
\begin{tabular}{lcccc|c}
\toprule
{} & Art & Clipart & Product & Real-World & Avg. \\ \hline
Baseline   & 58.71 & 44.20 & {71.75} & 73.19 & 61.96 \\
JiGen~\cite{li2017domain} & 53.04 & 47.51 & 71.47 & 72.79 & 61.20 \\
D-SAM~\cite{dinnocente2018domain}	& 58.03 & 44.37 & 69.22 & 71.45 & 60.77\\
DSON (Ours)       & \bf{59.37} & \bf{45.70} & \textbf{71.84} & \bf{74.68} & \bf{62.90} \\
\bottomrule
\end{tabular}}
\end{center}
\end{table}

\subsubsection{Datasets}

We evaluate the proposed method on three domain generalization benchmarks. 
The PACS dataset~\cite{pacs} is commonly used in domain generalization and is favored due to its large inter-domain shift across four domains: Photo, Art Painting, Cartoon, and Sketch. 
It contains a total of 9,991 images in 7 categories, with an image resolution of 227 $\times$ 227. 
We follow the experimental protocol in \cite{pacs}, where the model is trained on any three of the four domains (source domains), and then tested on the remaining domain (target domain). 
Office-Home~\cite{office-home} is a popular domain adaptation dataset, which consists of four distinct domains: Artistic Images, Clip Art, Product, and Real-world Images. 
Each domain contains 65 categories, with around 15,500 images in total. 
While the dataset is mostly used in the domain adaptation context, it can easily be repurposed for domain generalization by following the same protocol used in the PACS dataset. 
Finally, we employ five datasets---MNIST, MNIST-M, USPS, SVHN and Synthetic Digits--- for digit recognition and split training and testing subsets following \cite{xu2018deep}.
\subsubsection{Implementation details}
For the fair comparison with prior arts~\cite{balaji2018metareg,dinnocente2018domain,li2017domain,li2019episodic}, we employ ResNet as the backbone network in all experiments. 
The convolutional and BN layers are initialized with ImageNet pretrained weights. 
We use a batch size of 32 images per source domain, and optimize the network parameters over 10K iterations using SGD-M with a momentum 0.9 and an initial learning rate $\eta_0 = 0.02$. 
As suggested in~\cite{zhao2018adversarial}, the learning rate is annealed by $\eta_p = \frac{\eta_0}{(1+\alpha p)^\beta}$, where $\alpha = 10$, $\beta = 0.75$, and $p$ increases linearly from 0 to 1 as training progresses. We follow the domain generalization convention by training with the ``train'' split from each of the source domains, then testing on the combined ``train'' and ``validation'' splits of the target domain.

We made the mixture weights shared across all layers in our network to facilitate optimization.
In our experiments, the convergence rates of local mixture weights in lower layers were significantly slower than higher layers.
We sidestepped this issue by sharing the mixture weights across all the layers.
This strategy improved accuracy substantially and consistently in all settings.

\subsection{Comparison with Other Methods}

In this section, we compare our method with other domain generalization methods on PACS and Office-Home datasets.

\subsubsection{PACS}
Table~\ref{tab:PACS_result} portrays the domain generalization accuracy on the PACS dataset. 
The proposed algorithm is compared with several existing methods, which include JiGen~\cite{li2017domain}, D-SAM~\cite{dinnocente2018domain}, Epi-FCR~\cite{li2019episodic}, MetaReg~\cite{balaji2018metareg}, MASF~\cite{dou2019domain}, and MMLD~\cite{matsuura2020domain}.
Our method outperforms both the baseline and other state-of-the-art techniques by significant margins, which is particularly effective for {\it hard} domain (Sketch).
When ResNet-50 is employed as a backbone architecture, our method still achieves better performance than other baselines; this result  implies the scalability and stability of DSON when it is incorporated into a larger model.

\subsubsection{Office-Home}
We also evaluate DSON on the Office-Home dataset, and the results are presented in Table~\ref{tab:OH_result}. 
As in PACS, DSON outperforms the recently proposed JiGen~\cite{li2017domain} and D-SAM~\cite{dinnocente2018domain} as well as our baseline.
We find that DSON achieves the best score on all target domains.
Again, DSON is more advantageous in {\it hard} domain (Clipart).

\begin{table*}[!t]
	\begin{center}
	\caption{Domain generalization accuracy (\%) in ablation study. We compare DSON (ours) with its variants given by different implementations of normalization layers.
	`Art.' denotes `Art painting' domain in the PACS dataset.}
	\label{tab:ablation}
	\scalebox{0.9}{
		\setlength\tabcolsep{2pt}
		\begin{tabular}{lcccc|c|cccc|c}
			\toprule
			&\multicolumn{5}{c|}{PACS}&\multicolumn{5}{c}{Office-Home}\\
			{} & Art. & Cartoon & Sketch & Photo & Avg. & Art & Clipart & Product & Real-World & Avg.\\ \hline
			Baseline    & 78.47 & 70.41 & 70.68 & 95.87 & 78.86 & 58.71 & 44.20 & 71.75 & 73.19 & 61.96 \\
			IBN~\cite{ibn} & 75.29 & 72.95 & 77.42 & 92.04 & 79.43 & 55.41 & 44.82 & 68.28 & 71.95 & 60.09 \\
			DSBN~\cite{dsbn} & 78.61 & 66.17 & 70.15 & {95.51} & 77.61 & {59.04} & {45.02} & \textbf{72.67} & 71.98 & 62.18 \\
			SN~\cite{sn}          & {82.50} & {76.80} & {80.77} & {93.47} & {83.38} & 54.10 & 44.97 & {64.54} & {71.40} & {58.75}\\ 
			DSON (Ours)    & \textbf{84.67} & \textbf{77.65}  & \textbf{82.23} & \textbf{95.87} &  \textbf{85.11} & \textbf{59.37} & \textbf{45.70} & 71.84 & \textbf{74.68} & \textbf{62.90} \\
			\bottomrule
	\end{tabular}}
	\end{center}
 \end{table*}
 \begin{table*}[!t]
	\begin{center}
	\caption{Domain generalization accuracy (\%) on Digits datasets using a ResNet-18 architecture. We compare DSON (ours) with its variants given by different implementations of normalization layers.}
	\label{tab:digits_result}
	\scalebox{0.91}{
		\setlength\tabcolsep{7pt}
		\begin{tabular}{lccccc|cccc|c}
			\toprule
			&\multicolumn{5}{c|}{Digits}\\
			{} & MNIST & MNIST-M & USPS & SVHN & Synthetic & Avg. \\ \hline
			Baseline    & 86.15 & 74.44 & 90.07 & {\textbf{81.29}} & 94.46 & 85.28 \\
			DSBN~\cite{dsbn} & 87.01 & 71.20 & {91.18} & 78.23 & 94.30 & 84.38 \\
			SN~\cite{sn}     & {89.28} & {78.40} & {88.54} & {79.12} & \textbf{95.66} & {86.20} \\ 
			DSON (Ours)      & \textbf{89.62} & \textbf{79.00} & \textbf{91.63} & {81.02} & {95.34} & \textbf{87.32} \\
			\bottomrule
	\end{tabular}}
	\end{center}
\end{table*}

\subsection{Ablation Study}
\label{subsec:results}

\subsubsection{PACS and Office-Home Dataset} 
We conduct an ablation study to assess the contribution of individual components within our full algorithm on the PACS and Office-Home datasets.
Table~\ref{tab:ablation} presents the results, where our complete method is denoted by DSON.
It also presents accuracies of its variants given by different implementations of normalization layers.
We first present results from the baseline method, where the model is trained na\"ively with BN layers that are not specific to any single domain. 
Then, to examine the effects of domain-specific normalization layers, the BN layers are made specific to each of the source domains, which is denoted by DSBN~\cite{dsbn}. 
We also examine the suitability of SN~\cite{sn} by replacing BN layers with adaptive mixtures of BN, IN and LN.
IBN-Net~\cite{ibn} concatenates instance normalization and batch normalization in a channel axis, but its improvement is marginal.
We do not include batch-instance normalization (BIN)~\cite{bin} in our experiment because it is hard to optimize and the results are unstable.
The ablation study clearly illustrates the benefits of individual components in our algorithm: optimization of multiple normalization methods and domain-specific normalization.
Note that other normalization methods can degrade the performance compared to baseline depending on the dataset, while DSON consistently displays superior results.

\subsubsection{Digits Dataset}
The results on five digits datasets are shown in Table~\ref{tab:digits_result}. 
Our model achieves 87.32\% of average accuracy, outperforming all other baselines by large margins.

\begin{table}[t]
    \begin{center}
	\caption{Domain generalization accuracy (\%) using ResNet-18 in the presence of label noise on the PACS dataset. Note that $\Delta$Avg. denotes the amount of accuracy drop with respect to the results from clean data.}
	\label{tab:DG_noise}
	\scalebox{0.9}{
		\setlength\tabcolsep{6pt}
		\begin{tabular}{clcccc|c|c}
			\toprule
			 Noise level & Method & Art painting & Cartoon & Sketch & Photo & Avg. &$\Delta$Avg. \\ \hline
			 \multirow{5}{*}{0.2} & Baseline (BN) & 75.16 & 70.41 & 68.17 & 92.13 & 76.47 & -2.89 \\
			 & IBN~\cite{ibn} & 77.25 & 69.75 & 69.53 & 90.60 & 76.78 & -2.65\\
			 & DSBN~\cite{dsbn} & 77.10 & 66.00 & 59.43 & \textbf{94.85} & 74.35 & -3.27\\
			 & SN~\cite{sn} & 78.56 & {75.21} & 77.42 & 91.08 & 80.57 & -2.57 \\
			 & DSON (Ours) & \textbf{83.11} & \textbf{79.07} & \textbf{80.15} &  {94.79} & \textbf{84.03} & \textbf{-1.08} \\ \hline
			 \multirow{5}{*}{0.5} & Baseline (BN) & 73.49 & 64.68 & 58.95 & 89.22 & 71.59 & -7.77 \\
			 & IBN~\cite{ibn} & 67.60 & 63.58 & 65.08 & 86.53 & 70.70 & -8.73 \\
			 & DSBN~\cite{dsbn} & 75.05 & 57.98 & 59.99 & {93.35} & 71.59 & -6.02 \\
			 & SN~\cite{sn} & 78.27 & \textbf{74.06} & 72.23 & 88.86 & 78.36 & -4.78\\
			 & DSON (Ours) & \textbf{80.22} & {73.85} & \textbf{77.37} & \textbf{94.91} & \textbf{81.59} & \textbf{-3.52} \\
			\bottomrule
	\end{tabular}}
    \end{center}

\end{table}

\subsection{Additional Experiments}

\subsubsection{Robustness against Label Noise}
The performance of the proposed algorithm on the PACS dataset is tested in the presence of label noise, and the results are investigated against other approaches.
Two different noise levels are tested (0.2 and 0.5), and the results are presented in Table~\ref{tab:DG_noise}.
Although all algorithms undergo performance degradation, the amount of accuracy drops is marginal in general and DSON turns out to be more reliable with respect to label noise compared to other models.
This is partly because DSON makes the network less overfit to class discrimination, reducing the effects of noise.

\begin{table}[t]
	\begin{center}
	\caption{Multi-source domain adaptation results on the PACS dataset using \cite{shu2019transferable} as a backbone with a ResNet-18 architecture.}
	\label{tab:multi-source_DA}
	\scalebox{0.94}{
		\setlength\tabcolsep{8pt}
		\begin{tabular}{lcccc|c}
			\toprule
			{} & Art painting & Cartoon & Sketch & Photo & Avg. \\ \hline
			Baseline   & 78.85 & 76.79 & 78.14 & \textbf{99.42} & 83.30 \\
			DSBN~\cite{dsbn}	& \textbf{88.94} & {83.54} & 77.39 & \textbf{99.42} & {87.32}\\
			SN~\cite{sn}		& 79.00 & 77.66 & 79.11 & {97.78} & 83.39 \\
			DSON (Ours)      & {86.54} & \bf{88.61} & \textbf{86.93} & \textbf{99.42} & \textbf{90.38} \\
			\bottomrule
	\end{tabular}}
	\end{center}

\end{table}

\subsubsection{Multi-Source Domain Adaptation}
DSON can be extended to the multi-source domain adaptation task, where we gain access to unlabeled data from the target domain. 
{
To compare the effect of different normalization methods, we adopt the algorithm proposed by Shu \etal~\cite{shu2019transferable} as the baseline method and vary the normalization method only.}
The results are shown in Table~\ref{tab:multi-source_DA}, where we compare DSON with baseline, SN~\cite{sn}, and DSBN~\cite{dsbn}. 
All compared methods illustrate a large improvement over the baseline. 
In direct contrast to the results from the ablation analysis in Table~\ref{tab:ablation}, DSBN is clearly superior to SN. 
This is unsurprising, given that DSBN is focused specifically on the domain adaptation task. 
Interestingly, we find that DSON outperforms not only the baseline but also DSBN, which demonstrates how effectively DSON can be extended to the domain adaptation task.
Domain-specific models consistently outperform their domain-agnostic counterparts in this task.

\begin{figure}[t]
\begin{center}
	\subfigure[PACS]{{\includegraphics[width=0.45\linewidth]{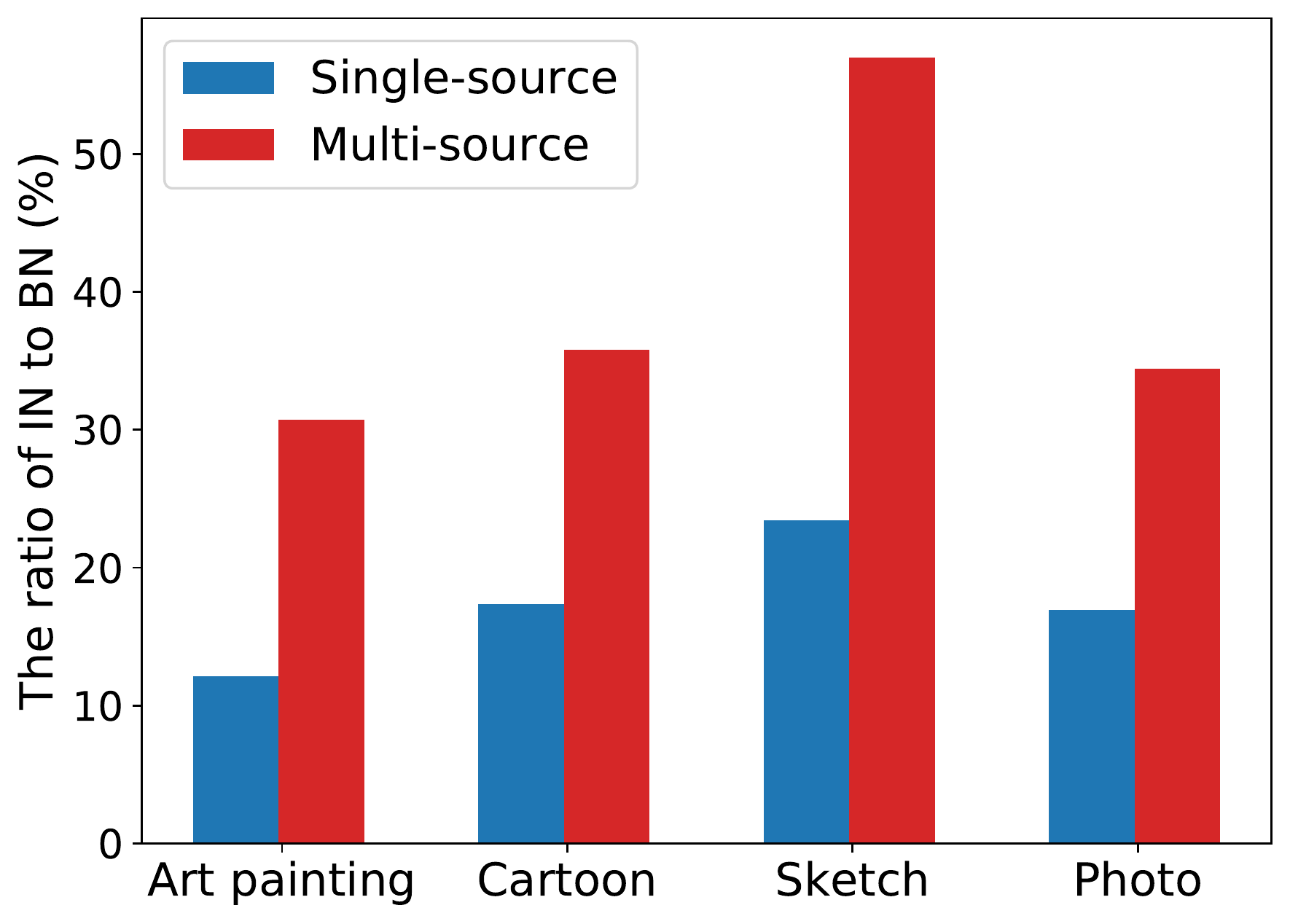}}}
	\hspace{0.3cm}
	\subfigure[Office-Home]{{\includegraphics[width=0.45\linewidth]{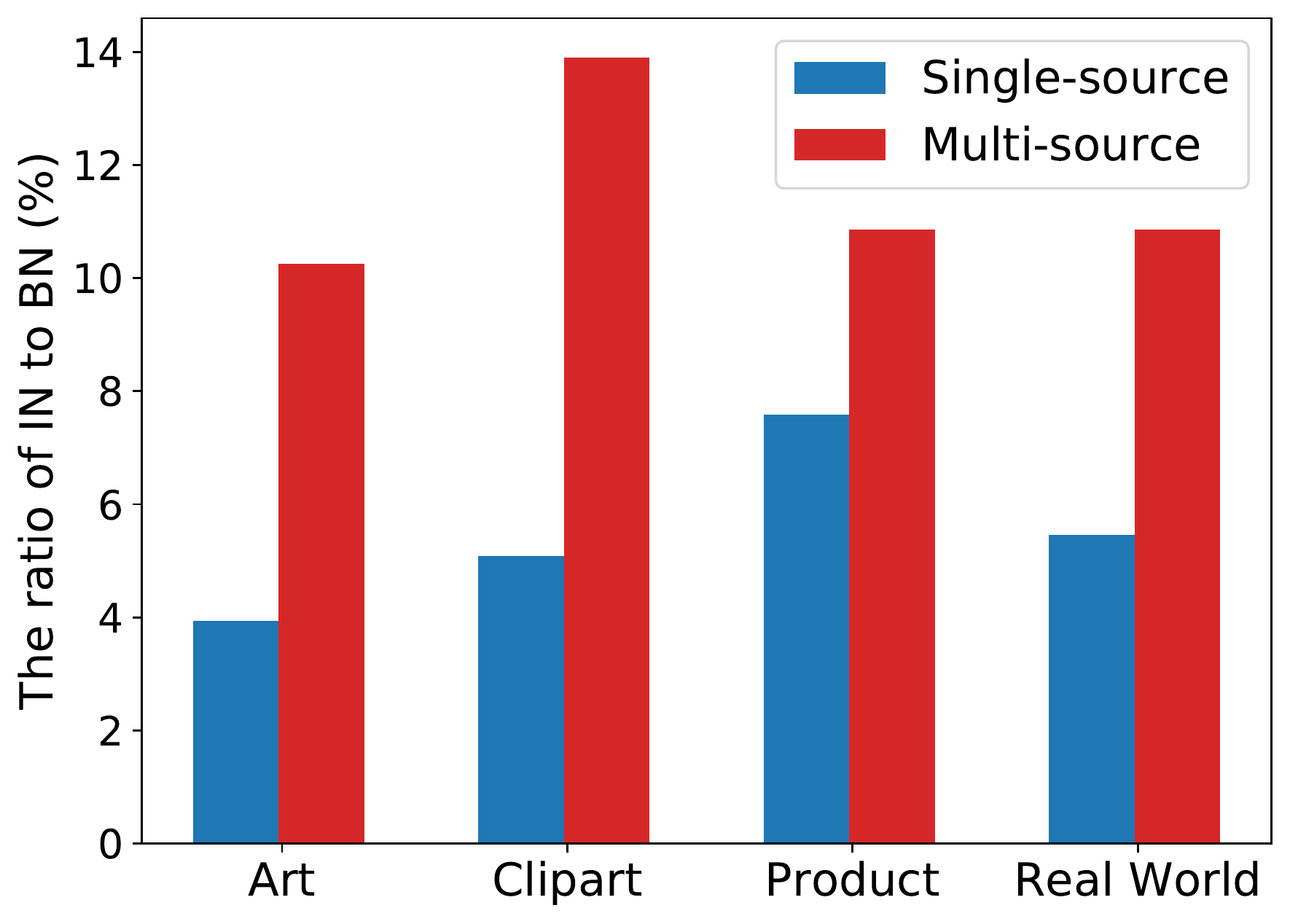}}}
	\caption{Analysis of the mixture weights with single-source and multi-source scenarios on the PACS (left) and Office-Home (right) datasets. 
	We present the weight ratio of IN to BN in our DSON module.
	(Best viewed in color.)}
	\label{fig:inbn}
\end{center}
\end{figure}

\subsection{Analysis}

\subsubsection{Mixture Weights} 
For a trained network with DSON in domain generalization tasks, the mixture weights of IN and BN are 3:7 and 1:9 on average for PACS and Office-Home datasets, respectively.
Additionally, we analyzed the effect that the number of source domains has on the value of the mixture weights.
To this end, we tested two scenarios; single-source and multi-source.
For each domain, single-source denotes that only the specified domain was used for training, while multi-source indicates that other domains, along with the specified domain, were used for training.
The graphs in Fig.~\ref{fig:inbn} show the weight ratio of IN to BN for each domain in the two scenarios.
As shown in the plot, training with multi-source domains led to a large and consistent increase in the usage of IN for all domains, because the multi-source scenario requires more domain-invariant representations than the single-source scenario.
We also validated the effectiveness of separating mixture weights for each domain independently in Table~\ref{tab:mixweight}, in which domain-specific mixture weights boosts the performance compared to the domain-agnostic ones.

\begin{table}[t]
	\begin{center}
	\caption{Effects of separating mixture weights for each domain in our DSON module on the PACS dataset with multi-source domain generalization scenario.
	Domain-agnostic denotes the mixture weights are shared across domains.}
	\label{tab:mixweight}
	\scalebox{0.94}{
		\setlength\tabcolsep{8pt}
		\begin{tabular}{l|cccc|c}
			\toprule
			Mixture weights & Art painting & Cartoon & Sketch & Photo & Avg. \\ \hline
			Domain-agnostic   & 82.13 & 74.10 & 80.02 & 95.03 & 82.82 \\
			Domain-specific (Ours)      & \textbf{84.67} & \textbf{77.65}  & \textbf{82.23} & \textbf{95.87} &  \textbf{85.11} \\
			\bottomrule
	\end{tabular}}
	\end{center}
 \end{table}
 \begin{table}[!t]
	\begin{center}
	\caption{Single-source domain generalization accuracy on the PACS dataset.
	Rows and columns denote source and target domains, respectively.
	We compare DSON (ours) with BN in terms of the amount of change (\%p).}
	\label{tab:singlesource}
	\scalebox{0.94}{
		\setlength\tabcolsep{4.5pt}
		\begin{tabular}{ccccc|cccc}
			\toprule
			&\multicolumn{4}{c|}{BN (\%)}&\multicolumn{4}{c}{DSON (\%p)}\\
			{} & A & C & S & P & A & C & S & P \\  \hline 
			Art painting & 89.42 & 60.71 & 48.74 & 94.25 & {\bf{--}} 1.20 & \bf{+ 7.42} & \bf{+ 20.36} & {\bf{--}} 0.44 \\
			Cartoon & 69.68 & 94.51 & 71.60 & 81.56 & \bf{+ 2.83} & {\bf{--}} 0.11 & \bf{+ 0.88} & \bf{+ 3.65}\\
			Sketch & 44.34 & 63.65 & 93.50 & 49.28 & \bf{+ 8.64} & \bf{+ 0.64} & {\bf{--}} 0.67 & \bf{+ 9.82}\\
			Photo & 61.72 & 29.10 & 33.98 & 97.86 & \bf{+ 3.56} & \bf{+ 16.85} & \bf{+ 1.19} & {\bf{--}} 0.52\\ 
			\bottomrule
	\end{tabular}}
	\end{center}
\end{table}

\subsubsection{Effects of Instance Normalization}
To analyze the effects of combining instance normalization, we tested single-source domain generalization on every source-target combination of domains, as shown in Table~\ref{tab:singlesource}.
Rows and columns denote source domains and target domains, respectively.
For evaluation, we used the validation split of the target domain in case source and target domains are the same.
Although DSON marginally sacrifices the accuracy compared to BN when source and target domains are the same, it brought a significant performance gain in most cross-domain scenarios. This presents a desirable trade-off in combining instance normalization; cross-category variance for domain invariance.

 \begin{table}[!t]
	\begin{center}
	\caption{Results from single source domain branch on the PACS dataset.
	Columns denote target domains and rows denote the single source domain branch of our model.
	}
	\label{tab:singlebranch}
	\scalebox{0.96}{
		\setlength\tabcolsep{8pt}
		\begin{tabular}{ccccc}
			\toprule
			{} & Art painting & Cartoon & Sketch & Photo \\  \hline 
			Art painting & - & 73.68 & 79.51 & 95.41 \\
			Cartoon & 80.96 & - & 75.87 & 93.77 \\
			Sketch & 78.71 & 71.72 & - & 91.98 \\
			Photo & 79.49 & 75.71 & 76.38 & - \\
			\hline
			DSON (all) & \bf{84.67} &\bf{77.65} & \bf{82.23} & \bf{95.87} \\
			\bottomrule
	\end{tabular}
	}
	\end{center}
\end{table}

\subsubsection{Results from Single Source Domain Branch}
We present the classification results using each single domain branch of our model on the PACS dataset in Table~\ref{tab:singlebranch}.
Columns denote target domains and rows denote each single source domain branch of our model.
It shows that the results from single domain branches differ slightly from each other in accuracy, and integrating them gives consistent performance gain.

%% file: sections/conclusion.tex

\section{Conclusion}
\label{sec:conclusion}

We presented a simple but effective domain generalization algorithm based on domain-specific optimized normalization layers.
The proposed algorithm uses multiple normalization methods while learning a separate affine parameter per domain.
The mixing weights are employed to compute the weighted average of multiple normalization statistics for each domain separately.
This strategy turns out to be helpful for learning domain-invariant representations since instance normalization removes domain-specific style while preserving semantic category information effectively.
The proposed algorithm achieves the state-of-the-art accuracy consistently on multiple standard benchmarks even with substantial label noise.
We showed that the algorithm is well-suited for unsupervised domain adaptation as well.
Finally we analyzed the characteristics and effects of our method with diverse ablative study.

\subsection*{Acknowledgement}

This work was supported by Institute for Information \&
Communications Technology Promotion (IITP) grant funded by the Korea government (MSIT) [2016-0-00563, 2017-0-01779].